\documentclass[11pt]{article}
\usepackage[final]{acl}
\usepackage{times}
\usepackage{latexsym}
\usepackage{amsmath,amssymb,amsthm}
\newtheorem{definition}{Definition}
\usepackage{booktabs}
\usepackage{multirow}
\usepackage{graphicx}
\usepackage{xcolor}
\usepackage{subcaption}
\usepackage{enumitem}
\usepackage[most]{tcolorbox}
\usepackage{float}

\newcommand{\IG}{\mathrm{IG}}

\title{When and How Should an Agent Clarify? \\ { CIGAsk: Teaching LLMs to Clarify via Counterfactual Information Gain}}

\author{
    Yunxiang Li$^{\heartsuit}$, Xixin Wu$^{\heartsuit}$, Helen Meng$^{\heartsuit}$ \\
    $^\heartsuit$The Chinese University of Hong Kong, Hong Kong SAR, China \\
    \texttt{yli@se.cuhk.edu.hk}
  }

\begin{document}
\maketitle

\begin{abstract}
Instruction-tuned LLMs faced with under-specified queries often commit to a single interpretation rather than ask for clarification, producing confidently wrong
  answers. In our experiments, prompting alone is insufficient: models either ask for clarification on every query or ask vague questions that fail to recover the
  missing information. Addressing this failure requires learning two coupled skills: \emph{when} to ask rather than answer and \emph{how} to ask a question that
  recovers the disambiguating information. Existing recipes either address only one of these skills or require a separately trained critic. We propose
  \textbf{CIGAsk}, an RL recipe that teaches both skills through two complementary reward signals within a multi-turn GRPO loop. \textbf{Counterfactual Information
  Gain (CIG)} compares the gold-answer log-likelihood under a frozen reference model with and without the user response, providing per-turn credit that guides
  \emph{how} to ask. The \textbf{Asymmetric Ambiguity Bonus} assigns a signed reward at the terminal token based on the gold ambiguity label, guiding \emph{when} to
  ask. Across three clarification benchmarks spanning table, passage, and open-domain QA, CIGAsk-7B outperforms the strongest external baseline despite using a
  smaller backbone. It also transfers across datasets without per-dataset tuning while preserving single-turn QA performance on out-of-distribution benchmarks.
\end{abstract}

\section{Introduction}
\label{sec:intro}

\begin{figure}[!t]
\centering
\includegraphics[width=\columnwidth]{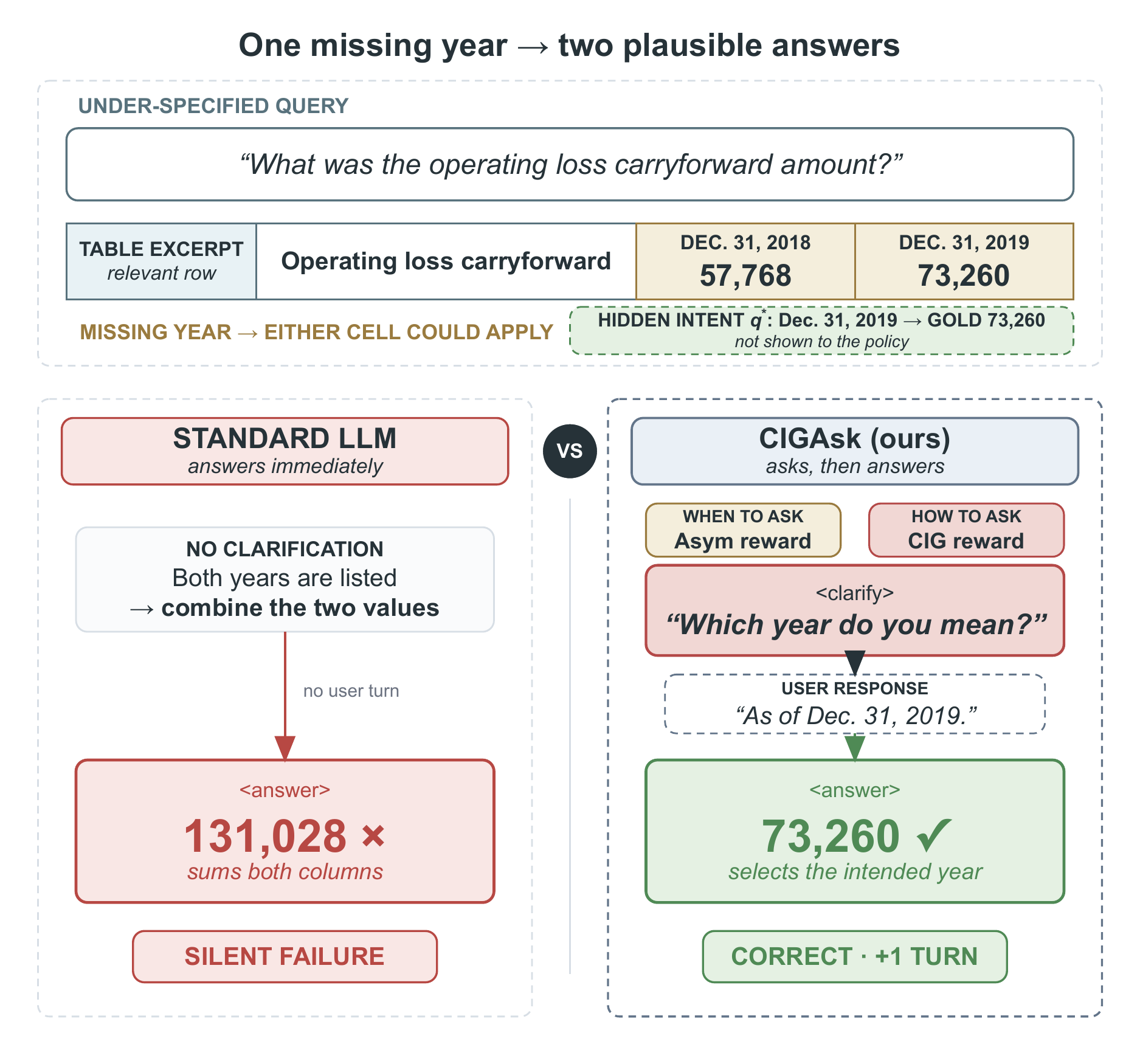}
\caption{\textbf{Clarification resolves a missing slot.}
  The underspecified query admits two plausible year-specific answers. A standard LLM answers immediately and combines them incorrectly, whereas CIGAsk learns when to
  clarify through the asymmetric reward and how to ask through CIG, recovering the intended year and the correct answer in one turn.}
\label{fig:teaser}
\end{figure}

\begin{figure*}[!t]
    \centering
    \includegraphics[width=\textwidth]{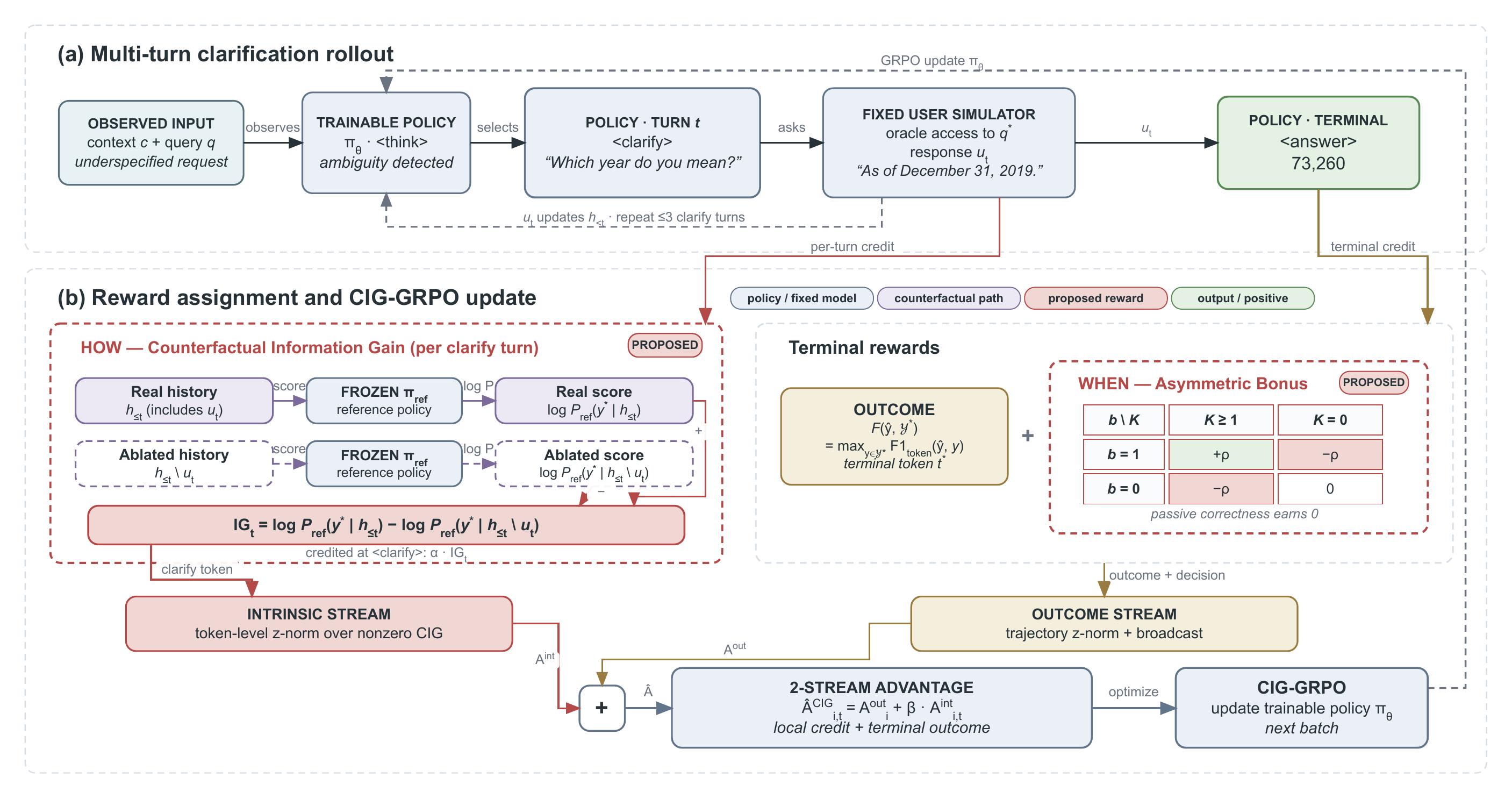}
\caption{\textbf{CIGAsk overview.}
  \textbf{(a)} A trainable policy $\pi_\theta$ interacts with a fixed user simulator with oracle access to the latent intent $q^*$, optionally requests clarification,
  and then produces a terminal answer.
  \textbf{(b)} CIG compares the gold-answer log-likelihood under the real history $h_{\leq t}$ and the response-ablated history $h_{\leq t}\setminus u_t$ using the
  same frozen reference policy $\pi_{\mathrm{ref}}$, providing local credit for \emph{how} to clarify. The outcome reward and Asymmetric Ambiguity Bonus provide
  terminal credit for answer quality and \emph{when} to clarify. After separate normalization, the intrinsic and outcome streams are combined into $\hat
  A^{\mathrm{CIG}}_{i,t}$ for the GRPO update.}
    \label{fig:method}
\end{figure*}

Instruction-tuned LLMs \citep{qwen25, deepseekr1} achieve strong single-turn performance on QA benchmarks. Yet when a request is under-specified, as in \emph{``What
  was the operating loss carryforward amount?''} for a table that lists both 2018 and 2019 figures \citep{deng2022pacific}, the same models often commit to one
  interpretation rather than ask for clarification, producing a plausible but potentially incorrect answer without acknowledging the ambiguity. The failure is easy to demonstrate, can
  be costly in deployment, and is hard to remove with prompting alone.

Addressing this failure requires the policy to learn two coupled skills in one training loop: \emph{when} to ask rather than answer, and \emph{how} to ask a
  question that recovers the disambiguating fact. Existing approaches address parts of this problem, but each has a limitation (\S\ref{sec:related}). SFT-based self-
  training \citep{andukuri2024stargate} filters trajectories by outcome reward and lacks a per-turn signal that grades how effectively a clarification elicits the
  missing fact. Offline action-preference learning \citep{chen2024act} cannot adapt to the policy's evolving trajectories. Rubric-guided RLVR \citep{askbench2026}
  needs per-instance rubric annotation that does not scale. Privileged-critic RL \citep{zhou2025sweetrl} requires privileged side information during training. None of
  these recipes combines a signal for \emph{when} to ask with per-turn credit for \emph{how} to ask using only the supervision already available in the training data.

 We close this gap with \textbf{CIGAsk}, a recipe that pairs two reward signals inside a multi-turn GRPO loop. \textbf{Counterfactual Information Gain (CIG)} grades
  each clarification by comparing the gold-answer likelihood under a \emph{frozen} reference model with and without the user response; the resulting log-ratio is the
  per-turn reward at the corresponding clarification token. Anchoring the score to a frozen model decouples the credit signal from the actor being optimized, so the
  reward does not drift as the actor changes, as it can in policy-coupled information-gain variants (\S\ref{sec:related}). The \textbf{Asymmetric Ambiguity Bonus}
  assigns a terminal bonus of $+\rho$, $-\rho$, or $0$ based on the ambiguity label and whether the rollout asks for clarification, with no positive bonus merely for
  not requesting clarification on a clear query. A token-level two-stream advantage carries both signals through GRPO and reduces to vanilla GRPO when no rollout
  requests clarification (\S\ref{sec:method:cig_grpo}).

\textbf{Our main contributions are as follows:}
\begin{itemize}[topsep=2pt,itemsep=2pt,leftmargin=12pt]
    \item We introduce \textbf{CIGAsk}, pairing a per-turn frozen-reference reward (CIG) that grades how well each clarification elicits the missing fact with a
    terminal asymmetric bonus that gates \emph{when} to ask.
  \item We show that CIGAsk improves clarification performance across three benchmarks and transfers across datasets without per-dataset tuning. It also preserves
  single-turn QA performance on out-of-distribution benchmarks without over-asking.
  \end{itemize}

\section{Related Work}
\label{sec:related}

\paragraph{Multi-turn clarification and ambiguous QA.}
Two threads frame the problem. Benchmarks establish the under-specification regime: AmbigQA \citep{min2020ambigqa}, AbgCoQA \citep{guo2021abgcoqa}, PACIFIC \citep{deng2022pacific}, ClariQ \citep{aliannejadi2020convai3}, QuestBench \citep{li2025questbench}; adjacent multi-answer disambiguation \citep{stelmakh2022asqa, lee2024ambigdocs, kim2023tree} treats the same ambiguity from the answer side rather than the question side. Question-generation methods predate LLM agents and span EVPI ranking \citep{rao2018learning}, conversational search \citep{aliannejadi2019asking, aliannejadi2020convai3}, and expected-information-gain criteria \citep{mazzaccara2024learning}; they target \emph{how} to ask while \emph{when} is assumed externally.

\paragraph{Training agents to clarify.}
Several lines of work train agents to clarify, differing in supervision regime. ACT \citep{chen2024act} pairs DPO-style action-level contrastive preferences with a Zephyr-7B base but does not score per-turn information value during training. AskBench \citep{askbench2026} uses rubric-guided RLVR; \citet{zhang2025futureturns} relabel RLHF preferences via simulated future turns but yield a binary signal; \citet{suri2025structureduncertainty} apply EVPI inside a fixed tool-call schema. Privileged-critic methods such as SWEET-RL \citep{zhou2025sweetrl} train a separately learned critic with ground-truth access at training time, but the critic is unavailable at deployment. Most similar in supervision is STaR-GATE \citep{andukuri2024stargate}, which self-trains on clarification trajectories filtered by outcome reward without per-turn credit; we differ precisely by the per-turn signal that decides \emph{which} clarification surfaced the disambiguating fact. Outcome-only multi-turn RL recipes such as UserRL \citep{salesforce2025userrl} treat clarification as an opaque sub-trajectory.

\paragraph{Per-turn credit assignment.}
Process reward models supply per-step credit but typically require step-labelled data: Math-Shepherd \citep{wang2024mathshepherd} synthesises labels via Monte Carlo, VinePPO \citep{kazemnejad2025vineppo} via rollout trees, ImplicitPRM \citep{yuan2025implicitprm} via outcome-derived logits. \citet{sullivan2025grpoprm} show vanilla GRPO with outcome reward equals a Monte-Carlo PRM only when group rollouts share identical prefixes, a condition multi-turn clarification rarely satisfies. Concurrent agentic variants assign turn-level credit via outcome propagation \citep{zeng2025turnlevel} or trajectory-DPO-derived implicit step rewards \citep{li2025istar}, but compute the signal under the training policy and inherit the same collapse risk. Closest to our signal are two concurrent per-turn IG recipes. IGPO \citep{wang2026igpo} computes turn-level rewards as the marginal increase in the \emph{training policy}'s probability of the correct answer for search agents; InfoPO \citep{kong2026infopo} computes the per-turn reward as the change in the \emph{agent}'s action distribution after user feedback for general user-centric agents. Neither uses a frozen reference: both score IG under the training policy itself, so the per-turn signal drifts as the actor updates, inheriting the degeneracy diagnosed by \citet{setlur2025scaling}. Neither pairs the per-turn signal with a label-driven gate on \emph{when} to ask. Our frozen-reference formulation (\S\ref{sec:method:reward}) decouples credit from the actor and targets proactive clarification specifically rather than search or generic collaboration. Information-theoretic and counterfactual rewards originate in active learning \citep{mackay1992information} and curiosity-driven exploration \citep{pathak2017curiosity}; our self-supervised process reward sits in this family but consumes only the outcome-level gold answer.

\section{Method}
\label{sec:method}

\subsection{Preliminaries}
\label{sec:method:prelim}

\paragraph{Multi-turn clarification setup.}
We formulate clarification as a POMDP with hidden state $q^*$ (the user's true intent). At turn $t$ the agent observes history $h_{<t}{=}\{(a_1,u_1),\ldots,(a_{t-1},u_{t-1})\}$, context $c$, and ambiguous question $q$, then picks an action from $\mathcal{A}{=}\{\textsc{Think},\textsc{Clarify},\textsc{Answer}\}$: \textsc{Think} runs internal reasoning; \textsc{Clarify} emits a question that triggers a user response $u_t\!\sim\!\text{Sim}(\cdot\mid q^*,c,h_{<t})$ from a fixed simulator; \textsc{Answer} emits $\hat y$ and terminates, after which $\hat y$ is scored against gold $y^*\!\in\!\mathcal Y^*$. The policy conditions on observable history rather than maintaining an explicit belief over $q^*$; the per-turn reward defined in \S\ref{sec:method:reward} supplies the supervision in its place.

\paragraph{GRPO.}
We train with Group Relative Policy Optimisation (GRPO, \citealp{shao2024deepseekmath}), a critic-free PPO variant. For each prompt $q$ we sample a group of $G$ output trajectories $\{o_i\}_{i=1}^{G}$ from the old policy $\pi_{\theta_{\text{old}}}$ with scalar returns $\{r_i\}_{i=1}^{G}$, compute the group-relative advantage $\hat A_{i,t}{=}\bigl(r_i{-}\mathrm{mean}(\{r_j\}_{j=1}^{G})\bigr)/\mathrm{std}(\{r_j\}_{j=1}^{G})$ broadcast to every token of $o_i$, and maximise the clipped PPO objective with a token-level KL penalty against a frozen reference $\pi_{\text{ref}}$:
\begin{align}
\mathcal J_{\text{GRPO}}(\theta) = \mathbb E_{q,\{o_i\}}\Bigl[ &\,\tfrac{1}{G}\!\sum_{i=1}^{G}\tfrac{1}{|o_i|}\!\sum_{t=1}^{|o_i|}\bigl\{L_{i,t}^{\text{PPO}}(\theta) \nonumber\\
& -\beta\,\mathbb D_{\text{KL}}[\pi_\theta\|\pi_{\text{ref}}]\bigr\}\Bigr], \nonumber\\
L_{i,t}^{\text{PPO}}(\theta) = \min\!\bigl(\rho_{i,t}\hat A_{i,t},\; &\,\mathrm{clip}_\epsilon(\rho_{i,t})\,\hat A_{i,t}\bigr),
\label{eq:grpo}
\end{align}
where $\rho_{i,t}{=}\pi_\theta(o_{i,t}\mid q,o_{i,<t})/\pi_{\theta_{\text{old}}}(o_{i,t}\mid q,o_{i,<t})$ is the per-token importance ratio and $\mathrm{clip}_\epsilon(\rho){=}\mathrm{clip}(\rho,1{-}\epsilon,1{+}\epsilon)$.

\subsection{Reward design}
\label{sec:method:reward}

\paragraph{Method overview.}
CIGAsk trains an LLM policy through multi-turn rollouts with a fixed user simulator with oracle access to the latent user intent (Figure~\ref{fig:method}). On top of the standard outcome reward (token-overlap F1 of the final answer), CIGAsk adds two reward terms placed at specific tokens:
\begin{align}
r_{i,t} =\; &\underbrace{F(\hat y_i, \mathcal Y^*)\,\mathbf{1}[t{=}t^*_i]}_{\text{outcome}} \nonumber\\
&+\underbrace{\alpha\,\IG_t\,\mathbf{1}[a_t{=}\textsc{Clarify}]}_{\text{CIG (per-turn)}} \nonumber\\
&+\underbrace{r^{\text{asym}}(b_i, K(\tau_i))\,\mathbf{1}[t{=}t^*_i]}_{\text{Asym (terminal)}},
\label{eq:total_reward}
\end{align}
  Here, $t_i^*$ denotes the terminal token, $\alpha$ is the CIG reward weight, and $K(\tau_i)$ indicates whether the policy requests clarification in trajectory $
  \tau_i$. Moreover, $F(\hat y, \mathcal Y^*){=}\max_{y\in\mathcal Y^*}\mathrm{F1}_{\text{token}}(\hat y, y)$ is the token-overlap F1 of the final answer; $\IG_t$ is
  the per-turn \textbf{Counterfactual Information Gain} scoring \emph{how} a clarification helps; and $r^{\text{asym}}$ is the trajectory-level \textbf{Asymmetric
  Ambiguity Bonus} scoring \emph{when} the policy chose to clarify, keyed on the gold ambiguity label $b_i$.

  \paragraph{Counterfactual Information Gain (CIG).}
  Supervised process-reward models require step-labelled supervision, which is unavailable for clarification (\S\ref{sec:related}). We instead measure value
  counterfactually: compare the real history $h_{\leq t}$ containing $u_t$ with the same history in which $u_t$ is ablated, and score the change in a frozen reference
  model's prediction of the gold answer.

  \begin{definition}[Outcome-derived process reward]
  \label{def:sspr}
  Let $\pi_{\mathrm{ref}}$ be the fixed reference policy and $P_{\mathrm{ref}}$ its induced answer distribution, $y^*$ the gold target, and $h_{\leq t}{=}h_{<t}\mathbin{\|}(a_t,u_t)$ a prefix containing $u_t$; write $h_{\leq t}\!
  \setminus\!u_t{=}h_{<t}\mathbin{\|}(a_t,\varnothing)$ for the same prefix with $u_t$ ablated, with context $c$ and question $q$ implicit in $h_{<t}$. The
  \emph{outcome-derived process reward} is the log-likelihood ratio
  \[
  r^{\mathrm{ssp}}_{\mathrm{ref}}(u_t; y^*\mid h_{<t})
  =\log\frac{P_{\mathrm{ref}}(y^*\mid h_{\leq t})}{P_{\mathrm{ref}}(y^*\mid h_{\leq t}\setminus u_t)}.
  \]
  \end{definition}

  \noindent CIG instantiates this reward for each \textsc{Clarify} action $a_t$:
  \begin{equation}
  \IG_t = \log P_{\mathrm{ref}}(y^*\mid h_{\leq t}) - \log P_{\mathrm{ref}}(y^*\mid h_{\leq t}\setminus u_t),
  \label{eq:cig}
  \end{equation}
  which we inject as $r^{\text{IG}}_t{=}\alpha\!\cdot\!\IG_t$ at the \textsc{Clarify} token, with $\alpha{=}0.3$ and $\IG_t$ clipped to $[-0.5,2.0]$ nats before
  scaling. The construction is related to predictive V-information \citep{xu2020theory}: CIG is a model-relative pointwise log-likelihood gain that measures how much
  observing $u_t$ changes the frozen reference model's support for $y^*$. It requires no separately learned critic and no per-step supervision beyond the dataset-
  level gold answer $y^*$ already used by the outcome reward.

\emph{Why frozen.} Scoring IG under the training policy ties the per-turn signal to the actor being optimized, allowing the reward to drift as the policy changes.
  \citet{setlur2025scaling} report a similar degeneracy for process-reward verifiers. \citet{sullivan2025grpoprm} show that vanilla GRPO with outcome reward is
  equivalent to a Monte Carlo PRM only when rollouts share prefixes, a condition rarely met in multi-turn clarification. Using a frozen predictor removes the reward's
  dependence on the evolving policy and does not require shared prefixes within the rollout group.

\paragraph{Asymmetric Ambiguity Bonus.}
  CIG provides credit only for clarification actions that appear in sampled rollouts; it provides no signal when a rollout contains no clarification. We therefore
  pair CIG with a trajectory-level bonus for \emph{when} to clarify. Let $K(\tau_i){=}\sum_t\mathbf{1}[a_t{=}\textsc{Clarify}]$ denote the number of clarification
  actions in rollout $\tau_i$, and let $b_i\!\in\!\{0,1\}$ be the gold ambiguity label. The bonus, assigned at the terminal token $t_i^*$, is
    \begin{equation}
    r^{\text{asym}}(b,K) = \begin{cases}
    +\rho & b{=}1,\;K{\geq}1 \\
    -\rho & b{=}1,\;K{=}0 \\
    -\rho & b{=}0,\;K{\geq}1 \\
    \phantom{+}0 & b{=}0,\;K{=}0,
    \end{cases}
    \label{eq:asym}
    \end{equation}
    The asymmetry lies in the last case: not requesting clarification on a clear query receives no ambiguity bonus rather than $+\rho$. A positive
  ambiguity bonus is reserved for trajectories that clarify ambiguous queries, while CIG separately scores how informative each clarification is. We use $\rho{=}0.30$ in the main experiments; sensitivity to the Asymmetric Ambiguity Bonus is reported in Table~\ref{tab:rho_sweep}.
    
 \paragraph{Advantage estimator.}
  \label{sec:method:cig_grpo}
  Vanilla GRPO assigns a single trajectory-level advantage to every token, so it does not preserve the locality of per-turn rewards. We use a two-stream variant that
  keeps each per-turn reward local. We split $r_{i,t}$ according to whether $t$ is the terminal-token index $t_i^*$ of $o_i$:
    \begin{equation}
    r^{\text{out}}_{i,t}{=}r_{i,t}\,\mathbf{1}[t{=}t^*_i],\quad r^{\text{int}}_{i,t}{=}r_{i,t}\,\mathbf{1}[t{\neq}t^*_i].
    \label{eq:split}
    \end{equation}
    The terminal stream $r^{\text{out}}$, which includes both the outcome reward and the asymmetric bonus, is normalised across trajectories as in GRPO and broadcast to
  every token in $o_i$. Let $\mu^{\text{out}}_G$ and $\sigma^{\text{out}}_G$ denote the group mean and standard deviation of $\{R^{\text{out}}_j\}_{j=1}^{G}$:
    \begin{equation}
    A^{\text{out}}_i = \frac{R^{\text{out}}_i - \mu^{\text{out}}_G}{\sigma^{\text{out}}_G+\epsilon},\qquad R^{\text{out}}_i{=}\sum_{t}r^{\text{out}}_{i,t}.
    \label{eq:a_out}
    \end{equation}
 The intrinsic stream is $z$-normalised at the token level over all non-zero $r^{\text{int}}_{i,t}$ values across rollouts in the same prompt group, with $
  \mu^{\text{int}}_G$ and $\sigma^{\text{int}}_G$ denoting their mean and standard deviation:
    \begin{equation}
    A^{\text{int}}_{i,t} = \frac{r^{\text{int}}_{i,t}-\mu^{\text{int}}_G}{\sigma^{\text{int}}_G+\epsilon}\,\mathbf{1}[r^{\text{int}}_{i,t}{\neq}0].
    \label{eq:a_int}
    \end{equation}
    The combined token-level advantage
    \begin{equation}
    \hat A^{\text{CIG}}_{i,t} = A^{\text{out}}_i + \beta\,A^{\text{int}}_{i,t},\quad \beta{=}1.0,
    \label{eq:cig_grpo}
    \end{equation}
    replaces $\hat A_{i,t}$ in Eq.~\ref{eq:grpo}. If all intrinsic rewards in the group are zero, we set $A^{\text{int}}_{i,t}{=}0$, and the estimator reduces to vanilla GRPO.
    
\section{Experiments}
\label{sec:experiments}

\subsection{Setup}
\label{sec:experiments:setup}

\paragraph{Models and datasets.}
We train Qwen2.5 \citep{qwen25} at 3B and 7B as the policy backbone, with GPT-4o \citep{openai2024gpt4o} as the user simulator with oracle access to the latent user intent (prompts in App.~\ref{app:training}). We evaluate on three clarification benchmarks spanning table-, passage-, and open-domain QA (Table~\ref{tab:datasets}). \textbf{PACIFIC} \citep{deng2022pacific} is the primary in-domain benchmark (table-grounded financial QA); \textbf{AbgCoQA} \citep{guo2021abgcoqa} tests cross-dataset recipe portability (same recipe re-trained, not zero-shot); \textbf{AmbigNQ} \citep{min2020ambigqa} adds an ambiguous-heavy open-domain regime. Single-turn retention checks appear in \S\ref{sec:analysis:ood}. And matched-backbone comparisons with ACT and SGP are reported in App.~\ref{app:matched_backbone}.

\begin{table}[H]
\centering
\footnotesize
\setlength{\tabcolsep}{6pt}
\caption{\textbf{Training datasets.} Three clarification benchmarks spanning table-, passage-, and open-domain QA.}
\label{tab:datasets}
\begin{tabular}{@{}lrrr@{}}
\toprule
\textbf{Dataset} & \textbf{Train} & \textbf{Val/Test} & \textbf{\%Ambig} \\
\midrule
PACIFIC  & 15{,}087 & 1{,}952 & 16\% \\
AbgCoQA  & 7{,}269  & 1{,}184 & 22\% \\
AmbigNQ  & 19{,}244 & 4{,}377 & 79\% \\
\bottomrule
\end{tabular}
\end{table}

\paragraph{Baselines.}
We compare three categories on the same Qwen2.5 backbone: (i) \textbf{Prompting-based}, \emph{Direct} (single-turn answer), \emph{FATA} (first-ask-then-answer with a fixed clarify template), and \emph{ReAct} \citep{yao2023react} (interleaved think--act prompt where the model chooses Clarify or Answer at each turn); (ii) \textbf{SFT-only}, our SFT warmstart without RL; and (iii) \textbf{External RL clarification recipes}, reported as context anchors rather than head-to-head: ACT \citep{chen2024act} (offline action-preference DPO on Zephyr-7B) and the concurrent SGP \citep{berant2026sgp} (cost-coefficient steerable self-play on Gemma-2-9b). Neither ACT nor SGP has fully released training code, so we cite their paper-reported numbers and leave cells we cannot fill under our protocol blank (Table~\ref{tab:main}).
\paragraph{Training protocol.}
For the main experiments, CIGAsk initialises from a supervised warmstart obtained by fine-tuning Qwen2.5 on curated multi-turn clarification trajectories
  derived from the benchmark training splits (details in App.~\ref{app:training}). RL training runs for up to 300 GRPO steps with 5 rollouts per prompt, sampling up
  to 3 clarify turns per rollout. The user simulator is a fixed GPT-4o instance prompted with the gold user intent $q^*$; it is never updated during training. We
  examine sensitivity to the simulator in Table~\ref{tab:simulator_sensitivity}. RL on the 7B backbone takes approximately 22 hours on $8\!\times\!$H100.

\paragraph{Metrics.}
On PACIFIC and AbgCoQA we report overall token-overlap F1 and post-clarify F1 ($\mathrm{F1}_{\mathrm{post}}$), the mean final-answer F1 among samples for which the
  policy issued at least one clarification; on AmbigNQ we report exact match (EM) following \citet{min2020ambigqa}. Across all three benchmarks we report clarify rates split by gold ambiguity label: Clr-amb (clarify rate on ambiguous samples, recall) and Clr-clr (clarify rate on clear samples, FPR). 
\subsection{Main results}
\label{sec:experiments:main}

\begin{table*}[!t]
\centering
\small
\setlength{\tabcolsep}{4pt}
\caption{\textbf{Main results.} \textbf{Bold} = our best model (CIGAsk 7B). PACIFIC fullval ($n{=}1{,}952$, GPT-4o sim); AbgCoQA / AmbigNQ $n{=}500$ seed-42 (GPT-4o sim). {---} = inapplicable or not reported. \textsuperscript{$\ddagger$}SGP / SGP-Oracle: Gemma-2-9b on 50/50 balanced dev; we report ambig/unambig F1 mean and place AmbigQA F1 in the AmbigNQ column.}
\label{tab:main}
\begin{tabular}{llrrrrrr}
\toprule
\multirow{2}{*}{\textbf{Method}} & \multirow{2}{*}{\textbf{Backbone}} & \multicolumn{4}{c}{\textbf{PACIFIC fullval} (1{,}952)} & \textbf{AbgCoQA} & \textbf{AmbigNQ} \\
\cmidrule(lr){3-6}
 & & \textbf{F1} & \textbf{F1}$_{\text{post}}$ & \textbf{Clr-amb} & \textbf{Clr-clr} & \textbf{F1} & \textbf{EM} \\
\midrule
\multicolumn{8}{l}{\textit{Prompting-based (no training)}} \\
Direct                       & Qwen2.5-7B & 0.312 & ---   & 0.000 & 0.000 & 0.320 & 0.058 \\
FATA                         & Qwen2.5-7B & 0.177 & 0.177 & 1.000 & 1.000 & 0.228 & 0.076 \\
ReAct \citep{yao2023react}   & Qwen2.5-7B & 0.481 & 0.418 & 0.329 & 0.410 & 0.554 & 0.076 \\
\midrule
\multicolumn{8}{l}{\textit{SFT only (no RL)}} \\
SFT                          & Qwen2.5-3B & 0.543 & ---   & 0.000 & 0.000 & 0.589 & 0.056 \\
SFT                          & Qwen2.5-7B & 0.581 & ---   & 0.000 & 0.000 & 0.638 & 0.109 \\
\midrule
\multicolumn{8}{l}{\textit{External RL clarification recipes (different base \& protocol; anchor only)}} \\
ACT \citep{chen2024act}                              & Zephyr-7B   & 0.681 & 0.620 & --- & --- & --- & --- \\
SGP \citep{berant2026sgp}\textsuperscript{$\ddagger$} & Gemma-2-9b  & 0.726 & --- & 0.429 & 0.206 & --- & 0.306 \\
SGP-Oracle\textsuperscript{$\ddagger$}               & Gemma-2-9b  & 0.787 & --- & 0.435 & 0.156 & --- & 0.359 \\
\midrule
\multicolumn{8}{l}{\textit{Ours (CIGAsk)}} \\
CIGAsk (3B)                  & Qwen2.5-3B & 0.673 & 0.633 & 0.539 & 0.179 & 0.635 & 0.271 \\
\textbf{CIGAsk (7B)}         & \textbf{Qwen2.5-7B} & \textbf{0.795} & \textbf{0.768} & \textbf{0.873} & \textbf{0.260} & \textbf{0.724} & \textbf{0.511} \\
\bottomrule
\end{tabular}
\end{table*}

CIGAsk establishes a new state of the art on PACIFIC fullval with a single Qwen2.5-7B policy, reaching F1 $0.795$ while also attaining the strongest reported post-
  clarify F1 and ambiguous-query clarification recall in Table~\ref{tab:main}. It surpasses the reported SGP and ACT results while maintaining a clear-query
  clarification rate well below FATA and comparable to SGP. Because the external results follow their original evaluation protocols, we provide matched-backbone
  comparisons in App.~\ref{app:matched_backbone}. Without access to gold ambiguity labels at evaluation time, CIGAsk jointly demonstrates high answer accuracy,
  productive clarification, and strong selectivity.

The baselines illustrate why prompting and SFT alone are insufficient. Direct commits to one interpretation of every query, while FATA forces every query through a
  clarification turn and consequently harms performance on clear inputs. ReAct lets the model choose whether to clarify but asks more often on clear than ambiguous
  queries, providing no positive selectivity. SFT improves answer quality but never clarifies, showing that the supervised warmstart alone does not learn when to ask.

\begin{figure}[H]
    \centering
    \includegraphics[width=\columnwidth]{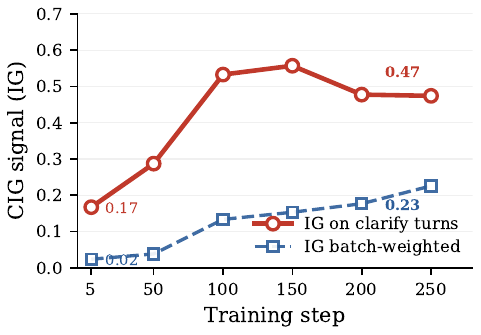}
\caption{\textbf{CIG grows during training} (CIGAsk 7B). Its rise coincides with the emergence of the recall--FPR gap.}
    \label{fig:cig_trajectory}
\end{figure}

CIGAsk 3B already outperforms every prompting and SFT baseline and nearly matches the reported ACT result with fewer than half as many parameters. Scaling from 3B
  to 7B improves both F1 and ambiguous-query clarification recall, while the clear-query false-positive rate increases only modestly. The resulting selectivity gap
  widens substantially, indicating that the larger model primarily learns to ask on ambiguous inputs rather than indiscriminately asking more often. The component
  ablation at 7B isolates the two reward terms (Table~\ref{tab:ablation}; \S\ref{sec:analysis:cig}): removing CIG lowers post-clarify F1, whereas removing the
  asymmetric bonus sharply reduces ambiguous-query recall. The asymmetric bonus therefore drives selectivity, which emerges during RL and is absent at the SFT
  warmstart (Fig.~\ref{fig:cig_trajectory}).

  The recipe transfers across datasets without per-dataset hyperparameter search. AbgCoQA differs from PACIFIC in grounding modality (passage vs.\ table) and dialogue
  structure (multi-turn conversational QA with history); when re-trained separately on AbgCoQA, the same recipe reaches F1 $0.724$ at 7B. AmbigNQ pushes the regime
  further: it is ambiguity-heavy, and uses exact match rather than token-overlap F1. CIGAsk 7B reaches EM $0.511$, substantially
  outperforming the prompting baselines and the 3B variant, indicating that the recipe benefits from increased backbone capacity in the harder open-domain setting.
  During training, CIG and the asymmetric bonus use only the gold answer $y^*$ and ambiguity label $b$ already provided by the benchmark.

\section{Analysis}
\label{sec:analysis}

\subsection{Does the Model Learn \emph{How} to Ask?}
\label{sec:analysis:cig}

CIG is the per-turn term that grades \emph{which} clarification recovers the disambiguating fact. We test whether it improves the usefulness of clarification beyond
  what outcome reward alone can shape. At 3B, CIGAsk and an outcome-only GRPO variant share the same SFT warmstart, seed, asymmetric bonus, KL, and rollout protocol,
  differing only in whether CIG is active. Both variants learn to clarify at similar rates, but only the CIG-equipped policy consistently turns those clarifications
  into improved post-clarify F1; the outcome-only variant falls to near-zero. At 7B, the effect is smaller but remains consistent: removing CIG reduces post-clarify
  F1 while leaving clarification selectivity largely intact (Table~\ref{tab:ablation}, $-$CIG). CIG therefore primarily improves the usefulness of the clarification
  rather than its frequency, complementing the asymmetric bonus that determines \emph{when} to ask. The larger effect at 3B suggests that smaller backbones benefit
  more from dense per-turn credit, consistent with prior observations on process rewards \citep{sullivan2025grpoprm}. The CIG signal strengthens over training as
  selective clarification emerges (Fig.~\ref{fig:cig_trajectory}). Across late-training checkpoints, items that the policy chooses to clarify also achieve higher F1
  than those answered directly, with bootstrap 95\% confidence intervals above zero throughout. This provides a self-consistency check that the learned policy directs
  clarification toward cases where interaction is productive.

\subsection{Does the Model Learn \emph{When} to Ask?}
\label{sec:analysis:asym}

A useful clarification policy must ask on ambiguous queries without over-asking on clear ones. We therefore measure selectivity by the gap between ambiguous-query
  recall and the false-positive rate on clear queries. Removing the asymmetric bonus while retaining CIG and the outcome reward sharply reduces recall without
  improving post-clarify F1 (Table~\ref{tab:ablation}, $-$asym). By contrast, removing CIG largely preserves selectivity but lowers post-clarify F1 ($-$CIG). These
  ablations separate the roles of the two components: the asymmetric bonus governs \emph{when} to ask, whereas CIG governs \emph{what} to ask and how useful the
  resulting clarification is. The asymmetric bonus also improves interaction quality: removing it roughly doubles the failure rate among generated clarification
  turns. The dominant failures are self-completion, where the model produces both a question and a plausible answer without yielding to the simulator, and declarative
  reformulation, where it restates the query rather than asking a genuine question. In both cases, the simulator receives no actionable request and therefore provides
  little disambiguating information. Without the decision-level bonus, the policy asks less often overall, and the questions it does ask fail more frequently.

\begin{table}[H]
\centering
\footnotesize
\setlength{\tabcolsep}{2.5pt}
\caption{\textbf{Component ablation on PACIFIC fullval.}}
\label{tab:ablation}
\begin{tabular}{@{}lcc rrrr@{}}
\toprule
\textbf{Variant} & \textbf{CIG} & \textbf{Asym} & \textbf{F1} & \textbf{F1\textsubscript{post}} & \textbf{Clr-amb} & \textbf{Clr-clr} \\
\midrule
\textbf{CIGAsk 7B} & \checkmark & \checkmark & \textbf{.795} & \textbf{.768} & \textbf{.873} & .260 \\
~~$-$asym        & \checkmark & ---        & .736 & .733 & .171 & .036 \\
~~$-$CIG         & ---        & \checkmark & .744 & .700 & .763 & .167 \\
~~$-$both        & --- & --- & .683 & .657 & .087 & .035 \\
\bottomrule
\end{tabular}
\end{table}
\subsection{Does the reference need to be frozen?}
  We isolate the effect of the frozen reference by replacing $\theta_0$ with the current actor when computing CIG, while keeping all other training and evaluation
  settings fixed. As shown in Table~\ref{tab:frozen_ref}, the current-actor variant underperforms the frozen-reference run on overall and post-clarification F1. It
  also clarifies fewer ambiguous queries and produces a substantially smaller selectivity gap. Its mean training reward changes little from early to late training.
  This pattern is consistent with the proposed mechanism: when the actor also scores its own clarifications, the counterfactual score changes as the policy is
  updated. Freezing the reference removes this coupling and provides a consistent basis for assigning clarification credit throughout training.

  \begin{table}[H]
  \centering
  \footnotesize
  \setlength{\tabcolsep}{2.5pt}
  \caption{\textbf{Frozen-reference ablation on PACIFIC fullval.}
  Reward reports mean training reward over steps 1--100 / 200--300.}
  \label{tab:frozen_ref}
  \begin{tabular}{@{}lrrrrr@{}}
  \toprule
  \textbf{CIG reference} & \textbf{F1} & \textbf{F1\textsubscript{post}} &
  \textbf{Clr-amb} & \textbf{Sel.} & \textbf{Reward} \\
  \midrule
  \textbf{Frozen (ours)}
  & \textbf{.795} & \textbf{.768} & \textbf{.873}
  & \textbf{+.613} & \textbf{.62/.78} \\
  Current actor
  & .607 & .359 & .233 & +.135 & .47/.53 \\
  \bottomrule
  \end{tabular}
  \end{table}
  
\subsection{Qualitative Case Study}
\label{sec:analysis:qualitative}

 A PACIFIC example illustrates the difference in clarification quality (Box~\ref{box:case}). The prompted baseline asks a generic question, eliciting only a
  restatement of the query before producing an incorrect aggregate. CIGAsk instead targets the missing year, recovers the relevant value in one turn, and answers
  correctly.

\begin{tcolorbox}[colback=gray!5,colframe=gray!55,arc=1.5mm,boxrule=0.5pt,left=5pt,right=5pt,top=4pt,bottom=4pt,fontupper=\footnotesize]
\refstepcounter{table}\label{box:case}\textbf{Box~\ref{box:case}: clarify quality dictates correctness.}\\[2pt]
\hrule\vspace{3pt}
\textbf{Context:} \emph{[table: deferred-tax-asset components by year, 2019 \& 2018]}\\
\textbf{Query:} \emph{What was the operating loss carryforward amount?}\\
\textbf{Intent:} \emph{\dots\ as of December~31, 2019}.\quad\textbf{Gold answer:} \texttt{73{,}260}\\[4pt]
\begin{minipage}[t]{0.475\columnwidth}
\centering\textbf{Prompted}~~$\boldsymbol{\times}$\\
\emph{generic clarify $\to$ wrong answer}\\[2pt]
\raggedright\footnotesize
\texttt{\#CLARIFY: Could you provide more context?}\\
\texttt{\#RESPOND: I'm asking about the operating loss carryforward.}\\
\texttt{\#ANSWER: 131{,}028}
\end{minipage}\hfill
\begin{minipage}[t]{0.475\columnwidth}
\centering\textbf{CIGAsk (ours)}~~\checkmark\\
\emph{targeted clarify $\to$ correct answer}\\[2pt]
\raggedright\footnotesize
\texttt{\#CLARIFY: Which period are you asking about?}\\
\texttt{\#RESPOND: as of December~31, 2019.}\\
\texttt{\#ANSWER: 73{,}260}
\end{minipage}
\end{tcolorbox}

\noindent This example highlights why deciding \emph{when} to clarify is not sufficient: the question must elicit information that resolves the ambiguity. CIG
  provides this content-level credit by rewarding clarifications whose responses move the frozen reference closer to the gold answer, encouraging targeted rather than
  generic questions.

\subsection{Out-of-Domain Retention and Selectivity}
\label{sec:analysis:ood}

Clarification-oriented reward shaping could encourage over-asking or weaken basic single-turn QA. We evaluate both risks on two open-QA benchmarks unseen during
  training: closed-book TriviaQA \citep{joshi2017triviaqa} and NQ Open \citep{kwiatkowski2019natural}. Under forced-direct evaluation, CIGAsk 7B performs comparably
  to the Qwen2.5-7B base on both benchmarks, showing that clarification training preserves standard QA ability (Table~\ref{tab:ood}). When given the multi-action
  prompt, the policy clarifies only a small fraction of queries, far below its rate on ambiguous PACIFIC inputs. Its questions target lexical ambiguity and temporal
  scope rather than defaulting to generic requests for more context (App.~\ref{app:clarify_examples}). These results show that CIGAsk transfers beyond PACIFIC's
  table-grounded setting while retaining a conservative and targeted clarification policy.

\begin{table}[H]
\centering
\footnotesize
\setlength{\tabcolsep}{4pt}
\caption{\textbf{OOD retention} on TriviaQA $+$ NQ Open closed-book ($n{=}500$, seed $42$).}
\label{tab:ood}
\begin{tabular}{@{}lrrr@{}}
\toprule
\textbf{Bench} & \textbf{CIGAsk F1} & \textbf{Base F1} & \textbf{Clarify-rate} \\
\midrule
TriviaQA & 56.2 & 54.0 & 5.6\% \\
NQ Open  & 25.9 & 25.1 & 9.4\% \\
\bottomrule
\end{tabular}
\end{table}

\section{Conclusion}
\label{sec:conclusion}

We introduced \textbf{CIGAsk}, a reinforcement learning recipe that teaches an instruction-tuned LLM both \emph{when} and \emph{how} to clarify under-specified
  queries within a multi-turn GRPO loop. CIGAsk combines two complementary signals: Counterfactual Information Gain provides per-turn credit for clarification
  quality, while the Asymmetric Ambiguity Bonus governs when clarification is warranted. Ablations show that CIG alone produces useful but infrequent clarifications,
  whereas the asymmetric bonus alone increases clarification recall but yields less productive questions. Their combination is therefore necessary for both selective
  and effective clarification.

  Empirically, CIGAsk 7B achieves state-of-the-art F1 on PACIFIC fullval ($0.795$). The same recipe, retrained without dataset-specific hyperparameter tuning,
  transfers to AbgCoQA and AmbigNQ while preserving single-turn QA ability on TriviaQA and NQ Open. More broadly, the frozen-reference counterfactual signal provides multi-turn credit assignment using only the gold answer already required by the outcome reward,
  without a learned critic or step-level supervision. The current asymmetric bonus still relies on gold ambiguity labels; for unannotated training data, a soft
  ambiguity signal estimated from sampled responses, such as semantic entropy, could reduce this dependence. Extending CIGAsk to label-free ambiguity estimation and
  to other agentic actions, such as tool use and search, is a promising direction.

 \paragraph{Limitations.}
  Our main experiments use a fixed GPT-4o simulator with oracle access to the latent user intent. Although a backbone-scale open simulator preserves the main result,
  the training setup still assumes a cooperative simulator that can respond accurately to clarification questions; robustness to more diverse, noisy, or human
  responses remains to be studied. The asymmetric bonus requires per-instance ambiguity labels during training, which may be unavailable in unannotated corpora.
  Future work could replace these labels with soft ambiguity estimates derived from sampled responses, such as semantic entropy. We validate the recipe across
  Qwen2.5, Zephyr, and Gemma backbones at the 3B--9B scale, but transfer to substantially larger models remains untested. Finally, all evaluated benchmarks are in
  English, leaving multilingual clarification as an open direction.

\paragraph{Ethical considerations.}
The asymmetric bonus inherits annotator judgements of what counts as ambiguous; downstream policies may behave unevenly across user populations whose phrasing conventions differ. Deployers should re-tune $\rho$ on their own distribution.

\section*{Acknowledgments}

This work is supported by Centre for Perceptual and Interactive Intelligence (CPII) Ltd, a
CUHK-led InnoCentre under InnoHK scheme of Innovation and Technology Commission.

\bibliography{references}

\appendix
\section{Implementation Details}
\label{app:training}

This appendix lists training hyperparameters (\S\ref{app:a1}), dataset sources and licenses (\S\ref{app:a2}), verbatim prompts (\S\ref{app:a3}), and qualitative clarify-turn examples on in-domain and out-of-domain probes (\S\ref{app:a4}).

\subsection{Training hyperparameters}
\label{app:a1}
Table~\ref{tab:hparams} lists every hyperparameter. The reward configuration is identical across PACIFIC and AbgCoQA; the AbgCoQA warmstart adds a small per-turn cost ($-0.03$) for long-horizon stability. All reported numbers are single-seed (PyTorch / vLLM seed $42$); the 3B paired comparison in Table~\ref{tab:ablation} is matched-seed across configurations.

\begin{table}[H]
\centering
\scriptsize
\setlength{\tabcolsep}{4pt}
\caption{\textbf{CIGAsk training hyperparameters.} Identical across PACIFIC and AbgCoQA.}
\label{tab:hparams}
\begin{tabular}{@{}llr@{}}
\toprule
\textbf{Group} & \textbf{Parameter} & \textbf{Value} \\
\midrule
\multirow{4}{*}{\textbf{Reward}}
& $\alpha$ (CIG weight) & 0.3 \\
& CIG clip range (nats) & $[-0.5, 2.0]$ \\
& $\rho$ (asym bonus) & 0.30 \\
& malformed-clarify penalty & $-0.30$ \\
\midrule
\multirow{6}{*}{\textbf{GRPO}}
& framework & verl \citep{verl2025} \\
& rollouts per prompt & 5 \\
& KL coefficient & 0.08 \\
& learning rate & $5{\times}10^{-7}$ \\
& steps (max) & 300 \\
& batch size & 32 \\
& $\epsilon_{\text{high}}/\epsilon_{\text{low}}$ & 1.4 \citep{yu2025dapo} \\
\midrule
\multirow{5}{*}{\textbf{Rollout}}
& engine & vLLM \citep{kwon2023vllm} \\
& temperature & 1.0 \\
& prompt tokens (max) & 6{,}144 \\
& response tokens (max) & 2{,}048 \\
& max clarify turns & 3 \\
\midrule
\multirow{4}{*}{\textbf{SFT warmstart}}
& engine & FSDP \citep{zhao2023fsdp} \\
& steps & 84 \\
& learning rate & $1{\times}10^{-5}$ \\
& corpus size & ${\sim}1{,}300$ multi-turn \\
\midrule
\multirow{4}{*}{\textbf{Compute}}
& 7B RL wall-clock & ${\sim}22$\,h \\
& 3B RL wall-clock & ${\sim}10$\,h \\
& SFT wall-clock & ${\sim}1.5$\,h \\
& hardware & 8$\times$H100 \\
\bottomrule
\end{tabular}
\end{table}

\subsection{Dataset details}
\label{app:a2}
Table~\ref{tab:dataset_details} lists size, split, prompting regime, source citation, and license for every dataset. PACIFIC, AbgCoQA, and AmbigNQ supply training and in-domain evaluation; TriviaQA and NQ Open are held-out probes for out-of-domain retention and the multi-action clarify probe (\S\ref{sec:analysis:ood}). All evaluations decode greedily ($T{=}0$ for direct extraction, $T{=}1.0$ for rollouts as in Table~\ref{tab:hparams}); scoring uses token-overlap F1 on PACIFIC and AbgCoQA and exact match on AmbigNQ following \citet{min2020ambigqa}, with regex extraction from the $\langle$answer$\rangle{\ldots}\langle/\text{answer}\rangle$ span.

\begin{table*}[!t]
\centering
\footnotesize
\setlength{\tabcolsep}{4pt}
\renewcommand{\arraystretch}{1.15}
\caption{\textbf{Dataset details.} Source, used size, prompting regime, and license for every dataset in the paper.}
\label{tab:dataset_details}
\begin{tabular}{@{}p{0.10\linewidth}p{0.30\linewidth}p{0.28\linewidth}p{0.16\linewidth}p{0.08\linewidth}@{}}
\toprule
\textbf{Dataset} & \textbf{Description} & \textbf{Used size \& protocol} & \textbf{Source} & \textbf{License} \\
\midrule
\multicolumn{5}{l}{\textit{Training and in-domain evaluation}} \\[2pt]
PACIFIC & Proactive conversational QA over financial tables and text; each example carries a gold ambiguity label and clarification reference. & Train $15{,}087$; fullval $1{,}952$ ($n_{\text{amb}}{=}316$, $n_{\text{cl}}{=}1{,}636$). Multi-turn with GPT-4o simulator; up to $3$ clarify turns. & \citet{deng2022pacific} & MIT \\
AbgCoQA & Conversational QA derived from CoQA with annotated ambiguous turns and clarification references. & Train $7{,}269$; fullval $1{,}184$; eval subset $n{=}500$ seed $42$ with GPT-4o simulator. & \citet{guo2021abgcoqa} & MIT \\
AmbigNQ & Ambiguous open-domain questions derived from Natural Questions, with multiple disambiguated targets per query. & Train $19{,}244$; test $4{,}377$; eval subset $n{=}500$ seed $42$ with GPT-4o simulator. & \citet{min2020ambigqa} & CC BY-SA 3.0 \\
\midrule
\multicolumn{5}{l}{\textit{Out-of-distribution probes (never trained on)}} \\[2pt]
TriviaQA (closed-book) & Trivia QA over short factoid questions; closed-book setting used as a single-turn retention check. & $n{=}500$ seed $42$; forced-direct and multi-action prompts (\S\ref{sec:analysis:ood}). & \citet{joshi2017triviaqa} & Apache 2.0 \\
NQ Open (closed-book) & Open-domain factoid questions from Natural Questions; closed-book setting. & $n{=}500$ seed $42$; same protocol as TriviaQA. & \citet{kwiatkowski2019natural} & Apache 2.0 \\
\bottomrule
\end{tabular}
\end{table*}

\subsection{Prompts}
\label{app:a3}
Verbatim system and user prompts used during training and evaluation.

\begin{tcolorbox}[colback=blue!4, colframe=blue!55!black, boxrule=0.6pt, arc=2pt,
                  left=8pt, right=8pt, top=6pt, bottom=6pt,
                  title=\textsc{Policy system prompt} (training \& eval),
                  fonttitle=\bfseries\footnotesize,
                  fontupper=\scriptsize\ttfamily, breakable]
You are a helpful agent. Available actions: THINK, CLARIFY, ANSWER.\\
- THINK: reason internally. Wrap in \textless think\textgreater\textless /think\textgreater.\\
- CLARIFY: ask ONE question. Wrap in \textless clarify\textgreater\textless /clarify\textgreater.\\
- ANSWER: give final answer. Wrap in \textless answer\textgreater\textless /answer\textgreater. ONLY the value, no explanation.\\
First THINK, then CLARIFY if ambiguous, then ANSWER.
\end{tcolorbox}

\begin{tcolorbox}[colback=violet!4, colframe=violet!60!black, boxrule=0.6pt, arc=2pt,
                  left=8pt, right=8pt, top=6pt, bottom=6pt,
                  title=\textsc{Simulator system prompt} (GPT-4o; PACIFIC),
                  fonttitle=\bfseries\footnotesize,
                  fontupper=\scriptsize\ttfamily, breakable]
You simulate a user who asked a question to a QA assistant.\\[2pt]
\textbf{ABSOLUTE RULES (violating any = failure):}\\
1. You do NOT know the answer. NEVER include numbers, dollar amounts, percentages, or computed values.\\
2. Answer ONLY the ONE dimension asked. If asked ``which year?'' $\to$ say ONLY the year. NEVER combine multiple pieces of info.\\
3. If the assistant's question is vague (e.g.\ ``more details?''), push back: ask THEM to be specific. Do NOT dump your info.\\
4. If the question is irrelevant, say so.\\
5. Keep response under 20 words.\\[2pt]
\textbf{GOOD examples.} \emph{Q: ``What is the revenue?'' $\mid$ Meant: 2019.} Assistant ``Which year?'' $\to$ ``Fiscal year 2019.'' Assistant ``Can you provide more details?'' $\to$ ``What specifically do you need?''\\[2pt]
\textbf{BAD examples.} ``From Q1 to Q2 in 2019, the gross profit increased.'' (combined period + metric + direction); ``The increase was \$13,407.'' (included a number); ``Sure, here are all the details $\ldots$'' (dumped everything).
\end{tcolorbox}

\begin{tcolorbox}[colback=gray!4, colframe=gray!60!black, boxrule=0.6pt, arc=2pt,
                  left=8pt, right=8pt, top=6pt, bottom=6pt,
                  title=\textsc{Per-turn user message} (each clarify turn),
                  fonttitle=\bfseries\footnotesize,
                  fontupper=\scriptsize\ttfamily, breakable]
Your original question: ``\textit{[ambiguous q]}''\\
What you actually meant: ``\textit{[gold intent}\,$q^*$\textit{]}''\\
The assistant asks: ``\textit{[clarification question]}''\\
How do you respond? (Brief, natural, NO numbers or data values)
\end{tcolorbox}

\subsection{OOD clarify-turn examples}
\label{app:a4}\label{app:clarify_examples}
On the small subset where CIGAsk 7B chooses to clarify rather than answer ($5.6\%$ on TriviaQA, $9.4\%$ on NQ Open; \S\ref{sec:analysis:ood}), the clarifications are sensible disambiguations: \emph{``Do you mean which US gangster was scheduled to be released from Alcatraz prison in November 1939?''} (year-specific Trivia query); \emph{``Which year are you asking about?''} on \emph{``who won oscar for best director this month''} (NQ query with a relative time reference); \emph{``Which episode are you asking about?''} on a series-by-name query that names the season but not the episode. The selectivity learned on PACIFIC table-grounded ambiguity transfers to text-grounded open QA without retraining.

\subsection{Sensitivity to the Asymmetric Ambiguity Bonus.}  
We vary the Asymmetric Ambiguity Bonus weight $\rho$ while holding all other settings fixed. Table~\ref{tab:rho_sweep} shows a non-monotonic effect. Among the values tested, $
  \rho{=}0.3$ gives the strongest overall F1, ambiguous-query recall, and selectivity. A lower weight provides a weaker incentive to clarify ambiguous queries,
  whereas increasing the weight beyond the main setting does not further improve selectivity and reduces both F1 and recall. We therefore use $\rho{=}0.3$ in the main
  experiments.

  \begin{table}[H]
  \centering
  \scriptsize
  \setlength{\tabcolsep}{5pt}
  \caption{\textbf{Sensitivity to the ambiguity-bonus weight on PACIFIC fullval.}
  }
  \label{tab:rho_sweep}
  \begin{tabular}{@{}lrrrr@{}}
  \toprule
  \textbf{Setting} & \textbf{F1} & \textbf{Clr-amb} &
  \textbf{Clr-clr} & \textbf{Sel.} \\
  \midrule
  $\rho{=}0.1$ & .702 & .681 & .175 & +.506 \\
  $\mathbf{\rho{=}0.3}$ \textbf{(main)}
                & \textbf{.795} & \textbf{.873} & \textbf{.260} & \textbf{+.613} \\
  $\rho{=}0.5$ & .703 & .789 & .188 & +.601 \\
  \bottomrule
  \end{tabular}
  \end{table}

  \subsection{Sensitivity to the user simulator.}
  CIG depends on the response produced by the user simulator. If that response is inconsistent with the latent intent, it can reduce or reverse the reward assigned to
  an otherwise useful clarification. We test this dependence by replacing GPT-4o with two fixed open-weight simulators, Qwen2.5-7B-Instruct and Qwen2.5-3B-Instruct
  \citep{qwen25}, while keeping all other settings fixed. Each simulator receives the same latent-intent information.

  \begin{table}[H]
  \centering
  \scriptsize
  \setlength{\tabcolsep}{3.5pt}
  \caption{\textbf{Sensitivity to the user simulator on PACIFIC fullval.}
  GPT-4o is the main setting.}
  \label{tab:simulator_sensitivity}
  \begin{tabular}{@{}lrrrr@{}}
  \toprule
  \textbf{Simulator} & \textbf{F1} & \textbf{F1\textsubscript{post}} &
  \textbf{Clr-amb} & \textbf{Sel.} \\
  \midrule
  GPT-4o (main)          & \textbf{.795} & \textbf{.768} & \textbf{.873} & +.613 \\
  Qwen2.5-7B-Instruct    & .783 & .752 & .856 & \textbf{+.636} \\
  Qwen2.5-3B-Instruct    & .677 & .536 & .462 & +.332 \\
  \bottomrule
  \end{tabular}
  \end{table}
The 7B simulator remains close to GPT-4o across answer quality and clarification recall, while achieving slightly higher selectivity. The 3B simulator produces
  substantially lower post-clarify F1 and ambiguous-query recall. This difference is consistent with simulator errors weakening the CIG signal: an intent-inconsistent
  response provides little useful evidence for the frozen reference, even when the policy asks a relevant question.

  In a manual audit of eight ambiguous PACIFIC cases, the 7B simulator returned an intent-consistent response in all eight cases, compared with one of eight for the
  3B simulator. Table~\ref{tab:simulator_examples} shows three representative examples.

  \begin{table}[H]
  \centering
  \scriptsize
  \setlength{\tabcolsep}{3pt}
  \renewcommand{\arraystretch}{1.12}
  \caption{\textbf{Representative simulator responses on ambiguous PACIFIC cases.}
  Three examples from the eight-case manual audit are shown.}
  \label{tab:simulator_examples}
  \begin{tabular}{@{}p{0.17\linewidth}p{0.16\linewidth}p{0.25\linewidth}p{0.30\linewidth}@{}}
  \toprule
  \textbf{Clarification} & \textbf{Latent intent} &
  \textbf{Qwen2.5-7B} & \textbf{Qwen2.5-3B} \\
  \midrule
  Which period?
  & Q1 to Q2
  & \emph{From Q1 to Q2} (\checkmark)
  & \emph{Fiscal Year 2019} ($\times$) \\

  Which expense?
  & SG\&A
  & \emph{SG\&A expenses} (\checkmark)
  & \emph{Research and development} ($\times$) \\

  Which period?
  & 2017 to 2018
  & \emph{From 2017 to 2018} (\checkmark)
  & \emph{2018; 2019} ($\times$) \\
  \bottomrule
  \end{tabular}
  \end{table}

  These results show that simulator quality can affect the CIG training signal, but GPT-4o is not necessary in this setting: the open-weight 7B simulator yields
  similar downstream performance.

\subsection{Matched-Backbone Comparisons}
  \label{app:matched_backbone}

  ACT \citep{chen2024act} and SGP \citep{berant2026sgp} use
  Zephyr-7B and Gemma-2-9B, respectively, whereas our main experiments use
  Qwen2.5. To reduce backbone confounding, we retrained CIGAsk on both external
  backbones using the same training recipe and without backbone-specific
  hyperparameter tuning. Table~\ref{tab:matched_backbone} compares these runs
  with the numbers reported by the corresponding methods.

\begin{table}[H]
    \centering
    \scriptsize
    \setlength{\tabcolsep}{3pt}
    \caption{\textbf{Matched-backbone comparisons on PACIFIC.}
    CIGAsk uses the same recipe across backbones. Reported baseline results
    follow their original evaluation protocols. Matching the backbone controls
    for model architecture, but not evaluation protocol.
    \textsuperscript{$\ddagger$} uses gold ambiguity labels.
    {---} denotes not reported.}
    \label{tab:matched_backbone}

    \begin{tabular}{@{}lrrrr@{}}
      \toprule
      \textbf{Method}
      & \textbf{F1}
      & \textbf{F1}$_{\text{post}}$
      & \textbf{Clr-amb}
      & \textbf{Clr-clr} \\
      \midrule

      \multicolumn{5}{@{}l}{\textit{Gemma-2-9B backbone}} \\
      SGP (reported)
      & 0.726 & --- & 0.429 & 0.206 \\
      SGP-Oracle\textsuperscript{$\ddagger$}
      & 0.787 & --- & 0.435 & 0.156 \\
      CIGAsk (ours)
      & 0.774 & 0.749 & \textbf{0.813} & \textbf{0.031} \\

      \addlinespace[2pt]

      \multicolumn{5}{@{}l}{\textit{Zephyr-7B backbone}} \\
      ACT (reported)
      & 0.681 & 0.620 & --- & --- \\
      CIGAsk (ours)
      & \textbf{0.709} & \textbf{0.686} & 0.712 & 0.044 \\

      \bottomrule
    \end{tabular}
  \end{table}
  On Gemma-2-9B, CIGAsk improves over the deployable SGP system in F1
  ($0.774$ vs.\ $0.726$), while remaining below the SGP-Oracle upper bound
  ($0.787$). It also exhibits substantially stronger selectivity than both SGP
  variants: CIGAsk clarifies on $81.3\%$ of ambiguous questions while
  over-clarifying on only $3.1\%$ of clear questions. On Zephyr-7B, CIGAsk
  exceeds ACT in both F1 ($0.709$ vs.\ $0.681$) and post-clarify F1
  ($0.686$ vs.\ $0.620$), while additionally providing explicit measurements
  of when the model chooses to clarify. Although the reported baselines follow
  their original evaluation protocols, the results show that CIGAsk's gains are
  not confined to the Qwen2.5 backbone and transfer across three model families.

\end{document}